\documentclass[10pt,twocolumn,letterpaper]{article}

\usepackage{cvpr}
\usepackage{times}
\usepackage{epsfig}
\usepackage{graphicx}
\usepackage{amsmath}
\usepackage{amssymb}

\usepackage{acronym}

\usepackage[breaklinks=true,bookmarks=false]{hyperref}

\cvprfinalcopy 

\def\cvprPaperID{10060} 

\newcommand{\inv}[1]{\ensuremath{#1}^{-1}}
\def\t2tk{t \rightarrow t+k}
\def\tk2t{t+k \rightarrow t}

\def\codelink{{\small \url{https://github.com/jspenmar/DeFeat-Net}}}

\graphicspath{{Images/}}

\acrodef{SSD}{Sum of Squared Differences}
\acrodef{FCN}{Fully Convolutional Network}
\acrodef{SPP}{Spatial Pooling Pyramid}
\acrodef{SfM}{Structure-from-Motion}
\acrodef{NMS}{Non-Maxima Suppression}
\acrodef{VO}{Visual Odometry}
\acrodef{SLAM}{Simultaneous Localization and Mapping}

\ifcvprfinal\fi
\begin{document}

\title{DeFeat-Net: General Monocular Depth via Simultaneous Unsupervised Representation Learning}

\author{Jaime Spencer,\quad Richard Bowden,\quad Simon Hadfield\\
Centre for Vision, Speech and Signal Processing (CVSSP)\\
University of Surrey\\
{\tt\small \{jaime.spencer, r.bowden, s.hadfield\}@surrey.ac.uk}}

\maketitle
\thispagestyle{empty}

\begin{abstract}
In the current monocular depth research, the dominant approach is to employ unsupervised training on large datasets, driven by warped photometric consistency. Such approaches lack robustness and are unable to generalize to challenging domains such as nighttime scenes or adverse weather conditions where assumptions about photometric consistency break down.

We propose DeFeat-Net (Depth \& Feature network), an approach to simultaneously learn a cross-domain dense feature representation, alongside a robust depth-estimation framework based on warped feature consistency. The resulting feature representation is learned in an unsupervised manner with no explicit ground-truth correspondences required.

We show that within a single domain, our technique is comparable to both the current state of the art in monocular depth estimation and supervised feature representation learning. However, by simultaneously learning features, depth and motion, our technique is able to generalize to challenging domains, allowing DeFeat-Net to outperform the current state-of-the-art with around 10\% reduction in all error measures on more challenging sequences such as nighttime driving.
\end{abstract}

\section{Introduction}
\begin{figure}[!t]
    \centering
    \includegraphics[width=0.49\linewidth,trim={0 35cm 0 0},clip]{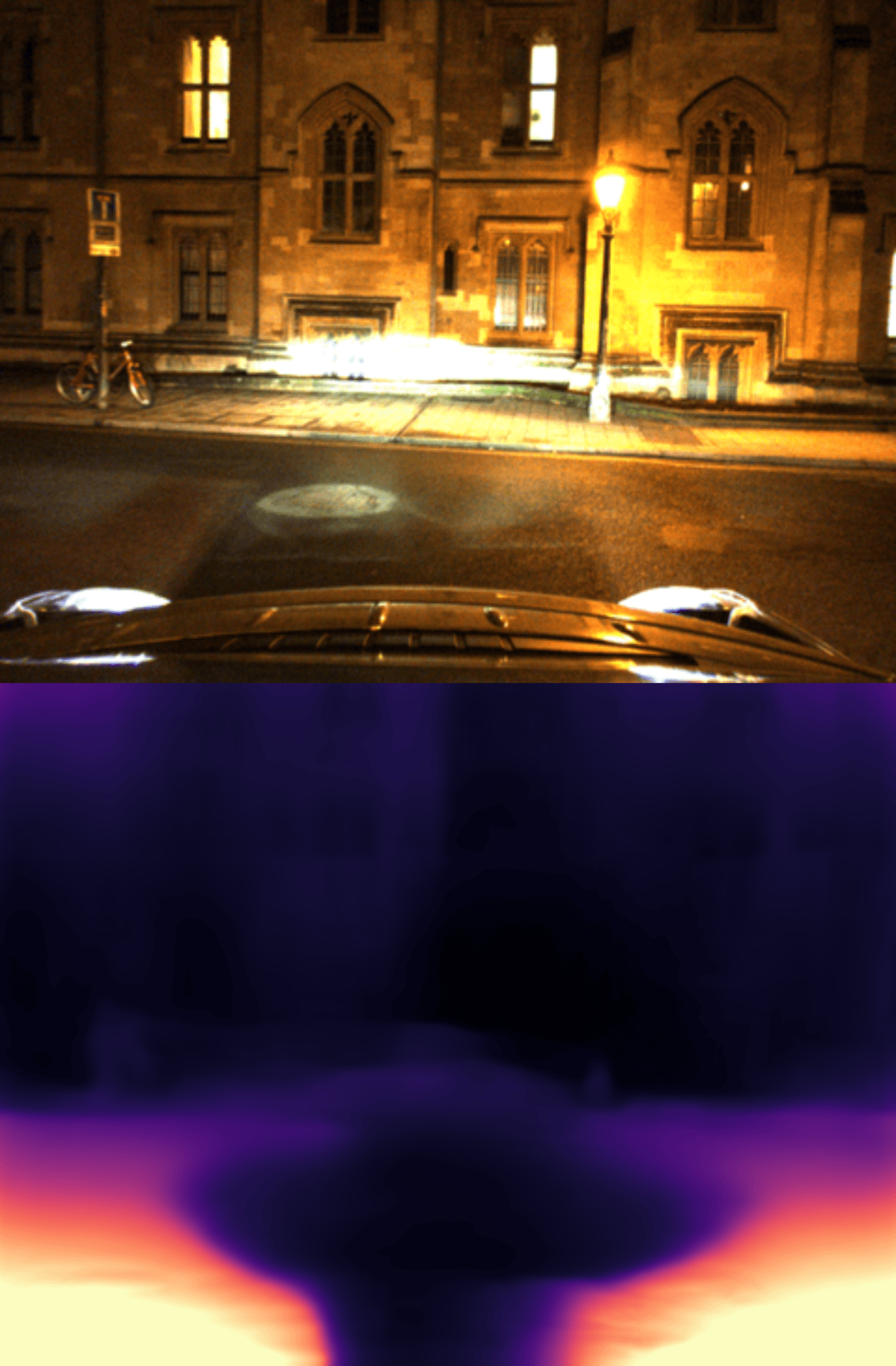}
    \includegraphics[width=0.49\linewidth,trim={0 0 0 35cm},clip]{baseline_night/000910}
    \caption{Left: Challenging lighting conditions during nighttime driving. Right: A catastrophic failure during depth map estimation for a current state-of-the-art monocular depth estimation framework, after being trained specifically for this scenario.}
    \label{fig:motivation}
\end{figure}

Recently there have been many advances in computer vision tasks related to autonomous vehicles, including monocular depth estimation \cite{Godard2019, Zhou2016, Wong2019} and feature learning \cite{Dusmanu2019, Schuster2019, Spencer2019}.
However, as shown in Figure~\ref{fig:motivation}, these approaches tend to fail in the most complex scenarios, namely adverse weather and nighttime conditions. 

In the case of depth estimation, this is usually due to the assumption of photometric consistency, which starts to break down in dimly-lit environments.
Feature learning can overcome such strong photometric assumptions, but these approaches tend to require ground truth pixel-wise correspondences and obtaining this ground truth in cross-seasonal situations is non-trivial.
Inconsistencies between GPS measurements and drift from \ac{VO} makes automatic pointcloud alignment highly inaccurate and manual annotation is costly and time-consuming.

We make the observation that depth estimation and feature representation are inherently complementary.
The process of estimating the depth for a scene also allows for the computation of ground-truth feature matches between any views of the scene. Meanwhile robust feature spaces are necessary in order to create reliable depth-estimation systems with invariance to lighting and appearance change.

Despite this relationship, all existing approaches tackle these challenges independently.
Instead, we propose DeFeat-Net, a system that is capable of jointly learning depth from a single image in addition to a dense feature representation of the world and ego-motion between consecutive frames.
What's more, this is achieved in an entirely self-supervised fashion, requiring no ground truth other than a monocular stream of images. 

We show how the proposed framework can use the existing relationships between these tasks to complement each other and boost performance in complex environments.
As has become commonplace \cite{Godard2017}, the predicted depth and ego-motion can be used to generate a correspondence map between consecutive images, allowing for the use of photometric error based losses. 
However, these correspondences can also be used as positive examples in relative metric learning losses \cite{Spencer2019}. 
In turn, the learnt features can provide a more robust loss in cases where photometric errors fail, \ie nighttime conditions. 

The remainder of the paper provides a more detailed description of the proposed DeFeat-Net framework in the context of previous work. 
We extensively show the benefits of our joint optimization approach, evaluating on a wide variety of datasets.
Finally, we discuss the current state-of-the-art and opportunities for future work.
The contributions of this paper can be summarized as:
\begin{enumerate}
	\item We introduce a framework capable of jointly and simultaneously learning monocular depth, dense feature representations and vehicle ego-motion.
	
	\item This is achieved entirely self-supervised, eliminating the need for costly and unreliable ground truth data collection.
	
	\item We show how the system provides robust depth and invariant features in all weather and lighting conditions, establishing new state-of-the-art performance.
\end{enumerate}
\section{Related Work}
Here we review some of the most relevant previous work, namely in depth estimation and feature learning. 

\subsection{Depth Estimation}
Traditionally, depth estimation relied on finding correspondences between every pixel in pairs of images.
However, if the images have been stereo rectified, the problem can be reduced to a search for the best match along a single row in the target image, known as disparity estimation. 
Initial methods for disparity estimation relied on hand-crafted matching techniques based on the \ac{SSD}, smoothness and energy minimization.

\textbf{Supervised.}
Ladick\`{y} \cite{Ladicky2015} and \v{Z}bontar \cite{Zbontar2015} showed how learning the matching function can drastically improve the performance of these systems. 
Mayer \etal \cite{Mayer2016} instead proposed DispNet, a \ac{FCN} \cite{Long2015} capable of directly predicting the disparity map between two images, which was further extended by \cite{Pang2017}. 
Kendall \etal \cite{Kendall2017} introduced GC-Net, where the disparity is processed as a matching cost-volume in a 3D convolutional network.
PSMNet \cite{Chang2018} and GA-Net \cite{Zhang2019} extended these cost-volume networks by introducing \ac{SPP} features and Local/Semi-Global aggregation layers, respectively.

Estimating depth from a single image seemed like an impossible task without these disparity and perspective cues.
However, Saxena \cite{Saxena2009} showed how it is possible to approximate the geometry of the world based on superpixel segmentation. Each superpixel's 3D position and orientation is estimated using a trained linear model and an MRF.
Liu \etal \cite{Liu2015a, Liu2015b} improve on this method by instead learning these models using a CNN, while Ladick\`{y} \etal \cite{Ladicky2014} incorporate semantic information as an alternative cue.

Eigen \etal \cite{Eigen2015, Eigen2014} introduced the first methods for monocular depth regression using end-to-end deep learning by using a scale-invariant loss.
Laina \cite{Laina2016} and Cao \cite{Cao2016} instead treated the task of monocular estimation as a classification problem and introduced a more robust loss function.
Meanwhile, Ummenhofer \etal \cite{Ummenhofer2017} introduced DeMoN, jointly training monocular depth and egomotion in order to perform \ac{SfM}. In this paper we go one step further, jointly learning depth, egomotion and the feature space used to support them.

\textbf{Unsupervised - Stereo Training.}
In order to circumvent the need for costly ground truth training data, an increasing number of approaches have been proposed using photometric warp errors as a substitute.
For instance, DeepStereo \cite{Flynn2016a} synthesizes novel views using raw pixels from arbitrary nearby views.
Deep3D \cite{Xie2016} also performs novel view synthesis, but restricts this to stereo pairs and introduces a novel image reconstruction loss.
Garg \cite{Garg2016} and Godard \cite{Godard2017} greatly improved the performance of these methods by introducing an additional autoencoder and left-right consistency losses, respectively. 
UnDeepVO \cite{Li2017a} additionally learns monocular \ac{VO} between consecutive frames by aligning the predicted depth pointclouds and enforcing consistency between both stereo streams.
More recently, there have been several approaches making use of GANs \cite{Aleotti2019, Pilzer2018}. 
Most notably, \cite{Sharma2019} uses GANs to perform day-night translation and provide an additional consistency to improve performance in nighttime conditions.
However, the lack of any explicit feature learning makes it challenging to generalize across domains.

\textbf{Unsupervised - Monocular Training.}
In order to learn unsupervised monocular depth without stereo information, it is necessary to learn a surrogate task that allows for the use of photometric warp losses. 
Zhou \etal \cite{Zhou2017a, Zhou2016} introduced some of the first methods to make use of \ac{VO} estimation to warp the previous and next frames to reconstruct the target view. 
Zhan \cite{Zhan2018} later extended this by additionally incorporating a feature based warp loss.
Babu \etal \cite{Babu2018a, Babu2018} proposed an unsupervised version of DeMoN \cite{Ummenhofer2017}.
Other published methods are based upon video processing with RNNs \cite{Wang2019a} and LSTMs \cite{Patraucean2016} or additionally predicting scene motion \cite{Vijayanarasimhan2017} or optical flow \cite{Janai2018, Wang2019b, Yin2018a}.

The current state-of-the-art has been pushed by methods that incorporate additional constraints \cite{Wang2018e} such as temporal \cite{Mahjourian2018}, semantic \cite{Chen2019}, edge \& normal \cite{Yang2018, Yang2018a}, cross-task \cite{Zou2018} and cycle \cite{Pilzer2019, Wong2019} consistencies.
Godard \etal \cite{Godard2019} expanded on these methods by incorporating information from the previous frame and using the minimum reprojection error in order to deal with occlusions.
They also introduce an automasking process which removes stationary pixels in the target frame. However, they still compute photometric losses in the original RGB colourspace, making it challenging to learn across domains.

\subsection{Feature Learning}
\textbf{Hand-Crafted.}
Initial approaches to feature description typically relied on heuristics based on intensity gradients in the image.
Since these were computationally expensive, it became necessary to introduce methods capable of finding interesting points in the image, \ie keypoints. 
Some of the most well-know methods include SIFT \cite{Lowe2004} and its variant RootSIFT \cite{Arandjelovic2012}, based on a Difference of Gaussians and \ac{NMS} for keypoint detection and HOG descriptors.

Research then focused on improving the speed of these systems. 
Such is the case with SURF \cite{Bay2006}, BRIEF \cite{Calonder} and BRISK \cite{Leutenegger2011}.
ORB features \cite{Rublee2012} improved the accuracy, robustness and speed of BRIEF \cite{Calonder} and are still widely used.

\textbf{Sparse Learning.}
Initial feature learning methods made use of decision trees \cite{Rosten2006}, convex optimization \cite{Simonyan} and evolutionary algorithms \cite{Krajnik2017, Krajnik2015} in order to improve detection reliability and discriminative power. Intelligent Cost functions \cite{Hadfield} took this a step further, by using Gaussian Processes to learn appropriate cost functions for optical/scene flow.

Since the widespread use of deep learning, several methods have been proposed to learn feature detection and/or description.
Balntas \etal \cite{Balntas} introduced a method for learning feature descriptors using in-triplet hard negative mining.
LIFT \cite{MooYi} proposes a sequential pipeline consisting of keypoint detection, orientation estimation and feature description, each performed by a separate network.
LF-Net \cite{Ono} builds on this idea, jointly generating dense score and orientation maps without requiring human supervision. 

On the other hand, several approaches make use of networks with shared encoder parameters in order to simultaneously learn feature detection and description. 
Georgakis \etal \cite{Georgakisa} learn 3D interest points using a shared Fast R-CNN \cite{Girshick2015} encoder.
Meanwhile, DeTone introduced SuperPoint \cite{Detone2018} where neither decoder has trainable parameters, improving the overall speed and computational cost. 
More recently, D2-Net \cite{Dusmanu2019} proposed a \textit{describe-then-detect} approach where the network produces a dense feature map, from which keypoints are detected using \ac{NMS}.

\textbf{Dense Learning.}
Even though SuperPoint \cite{Detone2018} and D2-Net \cite{Dusmanu2019} produce dense feature maps, they still focus on the detection of interest points and don't use their features in a dense manner.
Weerasekera \etal \cite{Weerasekera2019} learn dense features in the context of SLAM by minimizing multi-view matching cost-volumes, whereas \cite{Schonberger} use generative feature learning with scene completion as an auxiliary task to perform visual localisation.

The Universal Correspondence Network \cite{Choy2016} uses optical correspondences to create a pixel-wise version of the contrastive loss.
Schmidt \cite{Schmidt2017} instead propose semi-supervised training with correspondences obtained from KinectFusion \cite{Newcombe} and DynamicFusion \cite{Newcombea} models. 
Fathy \cite{Fathy2018a} and Spencer \cite{Spencer2019} extended the pixel-wise contrastive loss to multiple scale features through a coarse-to-fine network and spatial negative mining, respectively.
On the other hand, SDC-Net \cite{Schuster2019} focuses on the design of the network architecture, increasing the receptive field through stacked dilated convolution, and apply the learnt features to optical flow estimation.

In this work we attempt to unify state-of-the-art feature learning with monocular depth and odometry estimation. This is done in such a way that the pixel-wise correspondences from monocular depth estimation can support dense feature learning in the absence of ground-truth labels. Meanwhile, computing match-costs in the learned feature space greatly improves the robustness of the depth estimation in challenging cross-domain scenarios.
\section{Methodology}
\begin{figure*}[th!]
\vspace*{-1.5cm}
\centering
\includegraphics[width=\linewidth]{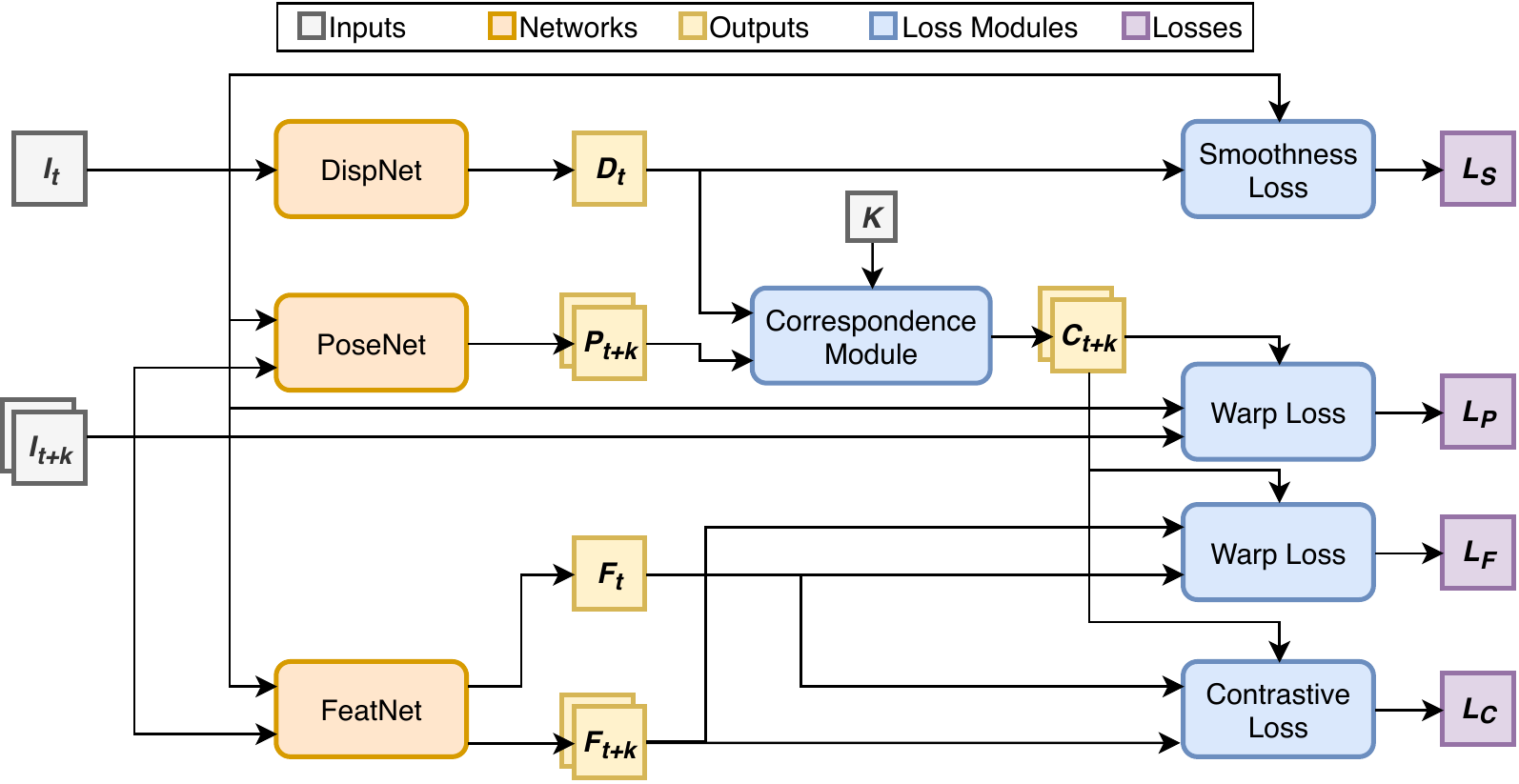}
\caption{Overview of \emph{DeFeat-Net} which combines complementary networks to simultaneously solve for feature representation, depth and ego-motion. The introduction of feature warping improves the robustness in complex scenarios.}
\label{fig: overview}
\end{figure*}
The main objective of DeFeat-Net is to jointly learn monocular depth and dense features in order to provide more robust estimates in adverse weather conditions.
By leveraging the synergy between both tasks we are able to do this in a fully self-supervised manner, requiring only a monocular stream of images.
Furthermore, as a byproduct of the training losses, the system additionally learns to predict \ac{VO} between consecutive frames.

Figure \ref{fig: overview} shows an overview of DeFeat-Net.
Each training sample is composed of a target frame $I_t$ and a set of support frames $I_{t+k}$, where $k \in \{ -1, 1 \}$.
Using the predicted depth for $I_t$ and the predicted transforms to $I_{t+k}$ we can obtain a series of correspondences between these images, which in turn can be used in the photometric warp and pixel-wise contrastive losses.
The code and pre-trained models for this technique will be available at \codelink.

\subsection{Networks}
\noindent
\textbf{DispNet.}
Given a single input image, $I_t$, its corresponding depth map is obtained through
\begin{equation} \label{eq: dispnet}
	D_t = \frac{1}{a\; \Phi_D(I_t) + b},
\end{equation}
where $a$ and $b$ scale the final depth to the range $[0.1, 100]$.
$\Phi_D$ represents the disparity estimation network, formed by a ResNet \cite{He2016} encoder and decoder with skip connections.
This decoder also produces intermediate disparity maps at each stage, resulting in four different scales.

\noindent
\textbf{PoseNet.}
Similarly, the pose prediction network $\Phi_P$ consists of a multi-image ResNet encoder, followed by a 4-layer convolutional decoder. 
Formally, 
\begin{equation} \label{eq: posenet}
	P_{\t2tk} = \Phi_P(I_t, I_{t+k}),
\end{equation}
where $P_{\t2tk}$ is the predicted transform between the cameras at times $t$ and $t+k$.
As in \cite{Godard2019, Wang2018e} the predicted pose is composed of a rotation in axis-angle representation and a translation vector, scaled by a factor of 0.001. 

\noindent
\textbf{FeatNet.}
The final network produces a dense $n$-dimensional feature map of the given input image, $\Phi_F: \mathbb{N}^{H \times W \times 3} \mapsto \mathbb{R}^{H \times W \times n}$.
As such, we define the corresponding L2-normalized feature map as 
\begin{equation}
	F = \left| \left| \Phi_F(I) \right| \right|.
\end{equation}
In this case, $\Phi_F$ is composed of a residual block encoder-decoder with skip connections, where the final encoder stage is made up of an \ac{SPP} \cite{Chang2018} with four scales.

\subsection{Correspondence Module}
Using the predicted $D_t$ and $P_{\t2tk}$ we can obtain a set of pixel-wise correspondences between the target frame and each of the support frames.
Given a 2D point in the image $p$ and its homogeneous coordinates $\dot{p}$ we can obtain its corresponding location $q$ in the 3D world through
\begin{equation}
	q = \inv{\pi} (\dot{p}) = \inv{K_t} \dot{p} \; D_t(p),
\end{equation}
where $\inv{\pi}$ is the backprojection function, $K_t$ is the camera's intrinsics and $D_t(p)$ the depth value at the 2D pixel location estimated using (\ref{eq: dispnet}). 

We can then compute the corresponding point $c_{\t2tk}$ by projecting the resulting 3D point onto a new image with
\begin{equation}
	c_{\t2tk}(p) = \pi(\dot{q}) = K_t P_{\t2tk} \dot{q},
\end{equation}
where $P_{\t2tk}$ is the transform to the new coordinate frame, \ie the next or previous camera position from (\ref{eq: posenet}).
Therefore, the final correspondences map is defined as 
\begin{equation}
    C_{\t2tk} = \left\{ c_{\t2tk}(p): \forall p  \right\}.
\end{equation}
These correspondences can now be used in order to determine the sampling locations for the photometric warp loss and the positive matches in a pixel-wise contrastive loss to learn an appropriate feature space.

\subsection{Losses}
Once again, it is worth noting that DeFeat-Net is entirely self-supervised. 
As such, the only ground truth inputs required are the orginal images and the camera's intrinsics.

\noindent
\textbf{Pixel-wise Contrastive.}
In order to train $\Phi_F$, we make use of the well established pixel-wise contrastive loss \cite{Choy2016, Schmidt2017, Spencer2019}.
Given two feature vectors from the dense feature maps, $f_1 = F_1(p_1)$ and $f_2 = F_2(p_2)$, the contrastive loss is defined as 
\begin{equation}
	l(y, f_1, f_2) = 
	\begin{cases}
	    \frac{1}{2}(d)^2 & \text{if } y = 1 \\
	    \frac{1}{2}\{\max(0, m - d)\}^2 & \text{if } y = 0 \\
    	0 & otherwise
    \end{cases} \vspace{-0.2cm}
    \label{eq:contrastive}
\end{equation}
with $y$ as the label indicating if the pair is a correspondence, $d = \left| \left| f_1 - f_2 \right| \right|$ and $m$ the target margin between negative pairs.  
In this case, the set of positive correspondences is given by $C_{\t2tk}$.
Meanwhile, the negative examples are generated using one of the spatial negative mining techniques from \cite{Spencer2019}.

From both sets, a label mask $Y$ is created indicating if each possible pair of pixels is a positive, negative or should be ignored. 
As such, the final loss is defined as 
\begin{equation}
	L_C = \sum_{p_1} \sum_{p_2} l( Y(p_1, p_2), F_t(p_1), F_{t+k}(p_2)).
\end{equation}
This loss serves to drive the learning of a dense feature space which enables matching regardless of weather and seasonal appearance variations.

\noindent
\textbf{Photometric and Feature Warp.}
We also use the correspondences in a differentiable bilinear sampler \cite{Jaderberg2015} in order to generate the warped support frames and feature maps
\begin{equation}
	I_{\tk2t} = I_{t+k} \langle C_{\t2tk} \rangle
\end{equation}
\begin{equation}
	F_{\tk2t} = F_{t+k} \langle C_{\t2tk} \rangle
\end{equation}
where $\langle \rangle$ is the sampling operator.
The final warp losses are a weighted combination of SSIM \cite{Wang2004} and L1, defined by
\begin{equation}
	\Psi(I_1, I_2) = \alpha \, \frac{1 \!-\! SSIM(I_1, I_2)}{2} + (1 \!-\! \alpha) \, \left| \left| I_1 \!-\! I_2 \right| \right|
\end{equation}
\begin{equation}
	L_P = \Psi(I_t, I_{\tk2t}),
\end{equation}
\begin{equation}
	L_F = \Psi(F_t, F_{\tk2t}),
\end{equation}
The photometric loss $L_P$ serves primarily to support the early stages of training when the feature space is still being learned.

\textbf{Smoothness.}
As an additional regularizing constraint, we incorporate a smoothness loss \cite{Heise2013}.
This enforces local smoothness in the predicted depths proportional to the strength of the edge in the original image, $\partial I_t$. 
This is defined as 
\begin{equation}
	L_S = \frac{\lambda}{N} \sum_{p}  \left| \partial D_t(p) \right| e^{- \left| \left| \partial I_t(p) \right| \right|},
\end{equation}
where $\lambda$ is a scaling factor typically set to 0.001.
This loss is designed to avoid smoothing over edges by reducing the weighting in areas of strong intensity gradients.
 
\subsection{Masking \& Filtering}
Some of the more recent improvements in monocular depth estimation have arisen from explicit edge-case handling \cite{Godard2019}.
This includes occlusion filtering and the masking of stationary pixels.
We apply these automatic procedures to the correspondences used to train both the depth and dense features. 

\noindent
\textbf{Minimum Reprojection.}
As the camera capturing the monocular stream moves throughout the scene, various elements will become occluded and disoccluded.
In terms of a photometric error based loss, this means that some of the correspondences generated by the system will be invalid. 
However, when multiple consecutive frames are being used, \ie $k \in \{ -1, 1 \}$, different occlusions occur in each image. 

By making the assumption that the photometric error will be greater in the case where an occlusion is present, we can filter these out by simply propagating the correspondence with the minimum error.
This is defined as 
\begin{equation}
    C_{\t2tk} \!=\! 
    \begin{cases}
        c_{t \rightarrow t-1} & \!\!\text{where } \Psi(I_t, I_{t \rightarrow t-1}) \!<\! \Psi(I_t, I_{t \rightarrow t+1})  \\
        c_{t \rightarrow t+1} & \!\!\text{otherwise}
    \end{cases}
\end{equation}

\noindent
\textbf{Automasking.}
Due to the nature of the training method and implicit depth priors (\ie regions further away  change less) stationary frames or moving objects can cause holes of infinite depth in the predicted depth maps.
An automasking procedure is used to remove these stationary pixels from contributing to the loss,
\begin{equation}
 \mu = \left[ \min_k \Psi(I_t, I_{t+k}) < \min_k \Psi(I_t, I_{\tk2t}) \right],
\end{equation}
where $\mu$ is the resulting mask indicating if a correspondence is valid or not and $[]$ is the Iverson bracket. In other words, pixels that exhibit lower photometric error to the unwarped frame than to the warped frame are masked from the cost function.
\begin{table*}[!t]
\vspace*{-1cm}
    \centering
    \begin{tabular}{c|ccccccc}
        Method & Abs-{Rel} & Sq-Rel & RMSE & RMSE-log & A1 & A2 & A3 \\ \hline
        LEGO \cite{Yang2018} & 0.162 & 1.352 & 6.276 & 0.252 & - & - & - \\
        Ranjan \cite{Ranjan2018} & 0.148 & 1.149 & 5.464 & 0.226 & 0.815 & 0.935 & 0.973 \\
EPC++ \cite{Luo2019} & 0.141 & 1.029 & 5.350 & 0.216 & 0.816 & 0.941 & 0.976 \\
Struct2depth ‘(M)’ \cite{Casser2019} & 0.141 & 1.026 & 5.291 & 0.215 & 0.816 & 0.945 & 0.979 \\
        Monodepth V2 \cite{Godard2019} & \textbf{0.123} & \underline{0.944} & \underline{5.061} & \textbf{0.197} & \textbf{0.866} & \textbf{0.957} & \textbf{0.980} \\
        \textbf{DeFeat} & \underline{0.126} & \textbf{0.925} & \textbf{5.035} & \underline{0.200} & \underline{0.862} & \underline{0.954} & \textbf{0.980} \\
    \end{tabular}
    \caption{Monocular depth evaluation on the KITTI dataset}
    \label{tab:kitti_depth}
\end{table*}

\begin{table*}[!t]
    \centering
    \begin{tabular}{c|ccccc}
    Method & $\mu_{+}$ & Global $\mu_{-}$ & Global AUC & Local $\mu_{-}$ & Local AUC \\ \hline
    ORB \cite{Rublee2012}  &  N/A & N/A & 85.83 & N/A & 84.06 \\
    ResNet \cite{He2016a} & 8.5117 & 25.9872 & 94.77 & 11.1335 & 68.26 \\
    ResNet-L2 & 0.341 & 1.0391 & 99.25 & 0.4371 & 71.80 \\
    VGG \cite{Simonyan2015a} & 4.0077 & 12.6543 & 92.94 & 5.9088 & 70.03 \\
    VGG-L2 & 0.3905 & 1.2235 & \underline{99.57} & 0.565 & 77.06 \\ 
    SAND-G \cite{Spencer2019} & \textbf{0.093} & 0.746 & \textbf{99.73} & 0.266 & 87.06 \\
    SAND-L  & 0.156 & 0.592 & 98.88 & 0.505 & \textbf{94.34} \\
    SAND-GL & 0.183 & 0.996 & 99.28 & 0.642 & \underline{93.34} \\
    \textbf{DeFeat} & \underline{0.105} & 1.113 & 99.10 & 0.294 & 83.64 \\
    \end{tabular}
    \caption{Learned feature evaluation on the KITTI dataset}
    \label{tab:kitti_feat}
\end{table*}

\section{Results}
Each subsystem in DeFeat-Net follows a U-Net structure with a ResNet18 encoder pretrained on ImageNet, followed by a 7 layer convolutional decoder similar to \cite{Godard2017}. 
The code and pre-trained models will be available at \codelink. In all our experiments, the warp loss parameter is set to  $\alpha = 0.85$ as per \cite{Jaderberg2015}.

On the KITTI dataset \cite{Geiger2012} we follow the Eigen-Zhou evaluation protocol of \cite{Godard2017, Zhou2016}. 
This dataset split provides 39,810 training images and 4,424 validation images. 
These images are all from a single domain (sunny daytime driving).

We also make use of the RobotCar Seasons dataset \cite{Sattler2018a}. 
This is a curated subset of the larger RobotCar dataset \cite{Maddern}, containing 49 sequences.
The dataset was specifically chosen to cover a wide variety of seasons and weather conditions, leading to greater diversity in appearance than KITTI.

Unlike the KITTI dataset, which provides sparse ground-truth depth from LiDAR, RobotCar Seasons does not include any depth ground truth. 
Our proposed technique is unsupervised, and can still be trained on this varied dataset, but the lack of ground truth makes quantitative evaluation on RobotCar Seasons impossible. 
To resolve this, we returned to the original RobotCar dataset and manualy created a validation dataset comprising of 12,000 images with their corresponding ground-truth LiDAR depth maps, split evenly across day and night driving scenarios.

\begin{table*}[!btp]
\vspace*{-1cm}
    \centering
    \begin{tabular}{c|c|ccccccc}
        Test domain & Method & Abs-{Rel} & Sq-Rel & RMSE & RMSE-log & A1 & A2 & A3 \\ \hline
Day & Monodepth V2 \cite{Godard2019} & 0.271 & 3.438 & 9.268 & 0.329 & \textbf{0.600} & 0.840 & 0.932\\
Day & \textbf{DeFeat} & \textbf{0.265} & \textbf{3.129} & \textbf{8.954} & \textbf{0.323} & 0.597 & \textbf{0.843} & \textbf{0.935} \\ \hline
Night & Monodepth V2 \cite{Godard2019} & 0.367 & 4.512 & 9.270 & 0.412 & 0.561 & 0.790 & 0.888\\ 
Night & \textbf{DeFeat} & \textbf{0.335} & \textbf{4.339} & \textbf{9.111} & \textbf{0.389} & \textbf{0.603} & \textbf{0.828} & \textbf{0.914 }\\
    \end{tabular}
    \caption{Monocular depth evaluation on the RobotCar dataset}
    \label{tab:robotcar_depth}
\end{table*}

\begin{figure*}[!t]
    \centering
    \renewcommand{\tabcolsep}{0pt}
    \begin{tabular}{cccc}
    \includegraphics[width=0.24\linewidth,trim={0 3cm 0 0},clip]{baseline_night/000910}
    &\includegraphics[width=0.24\linewidth,trim={0 3cm 0 0},clip]{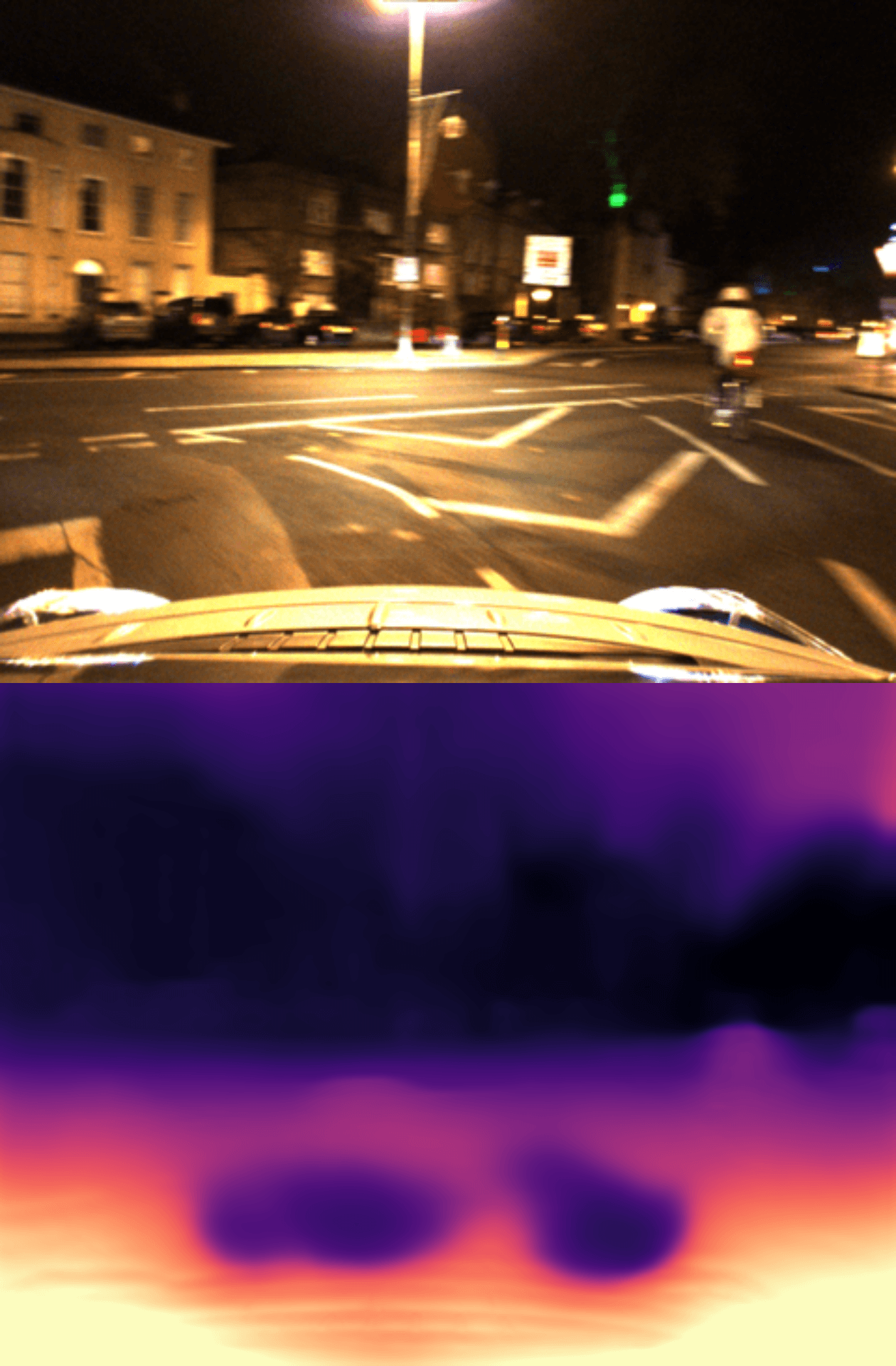}
    &\includegraphics[width=0.24\linewidth,trim={0 3cm 0 0},clip]{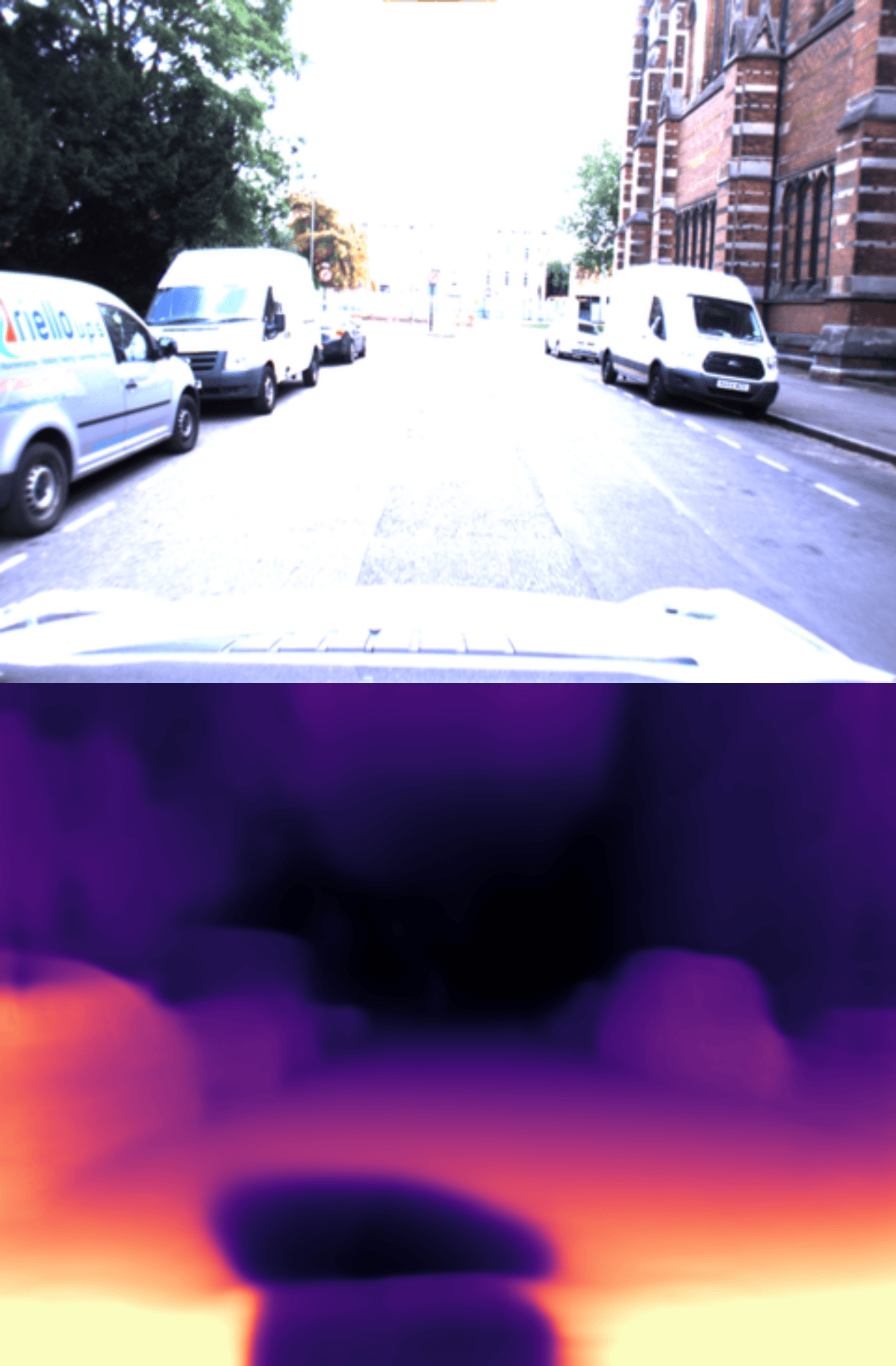}
    &\includegraphics[width=0.24\linewidth,trim={0 3cm 0 0},clip]{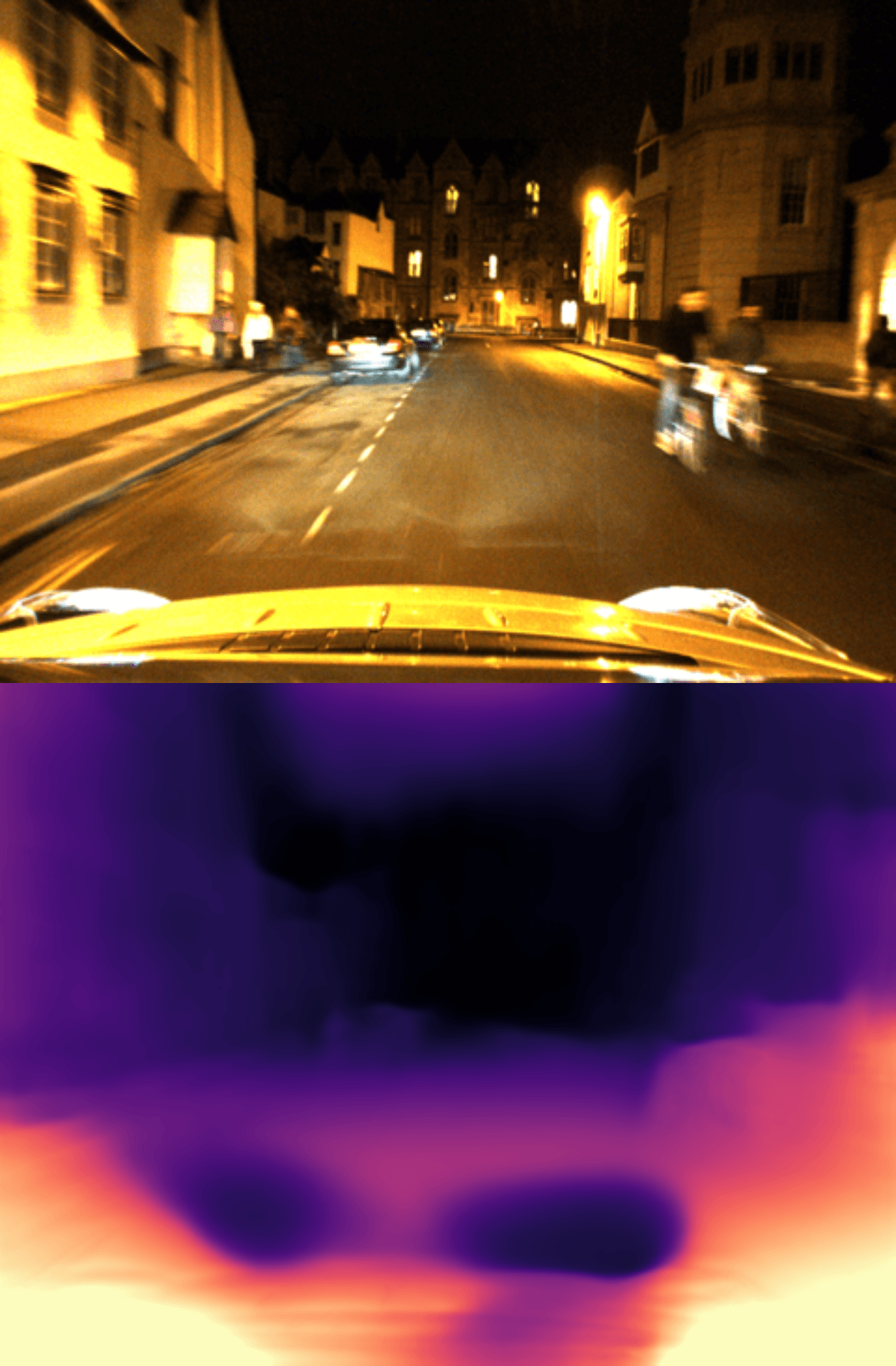} 
    \\
    \includegraphics[width=0.24\linewidth,trim={0 3cm 0 35cm},clip]{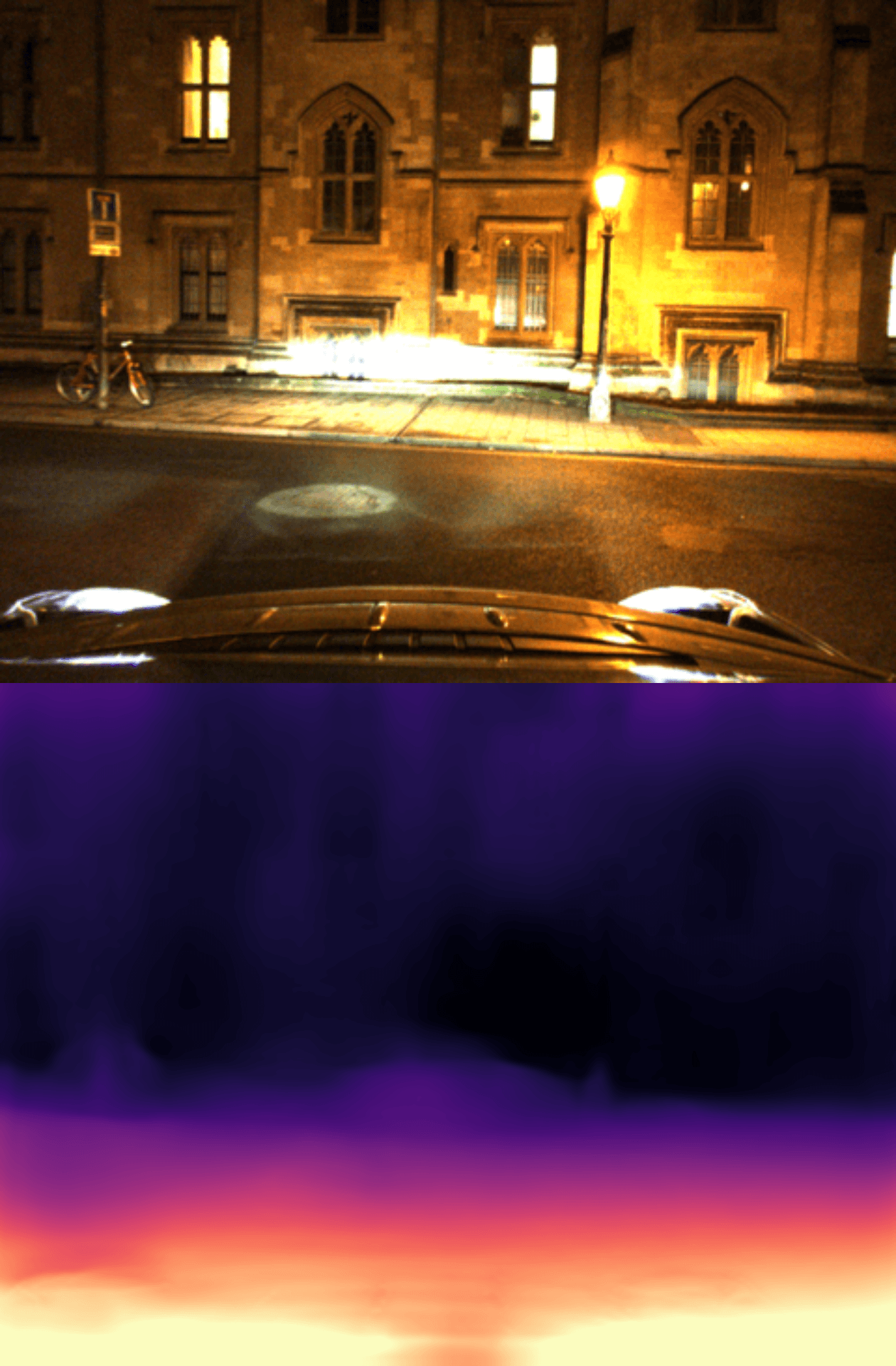} 
    &\includegraphics[width=0.24\linewidth,trim={0 3cm 0 35cm},clip]{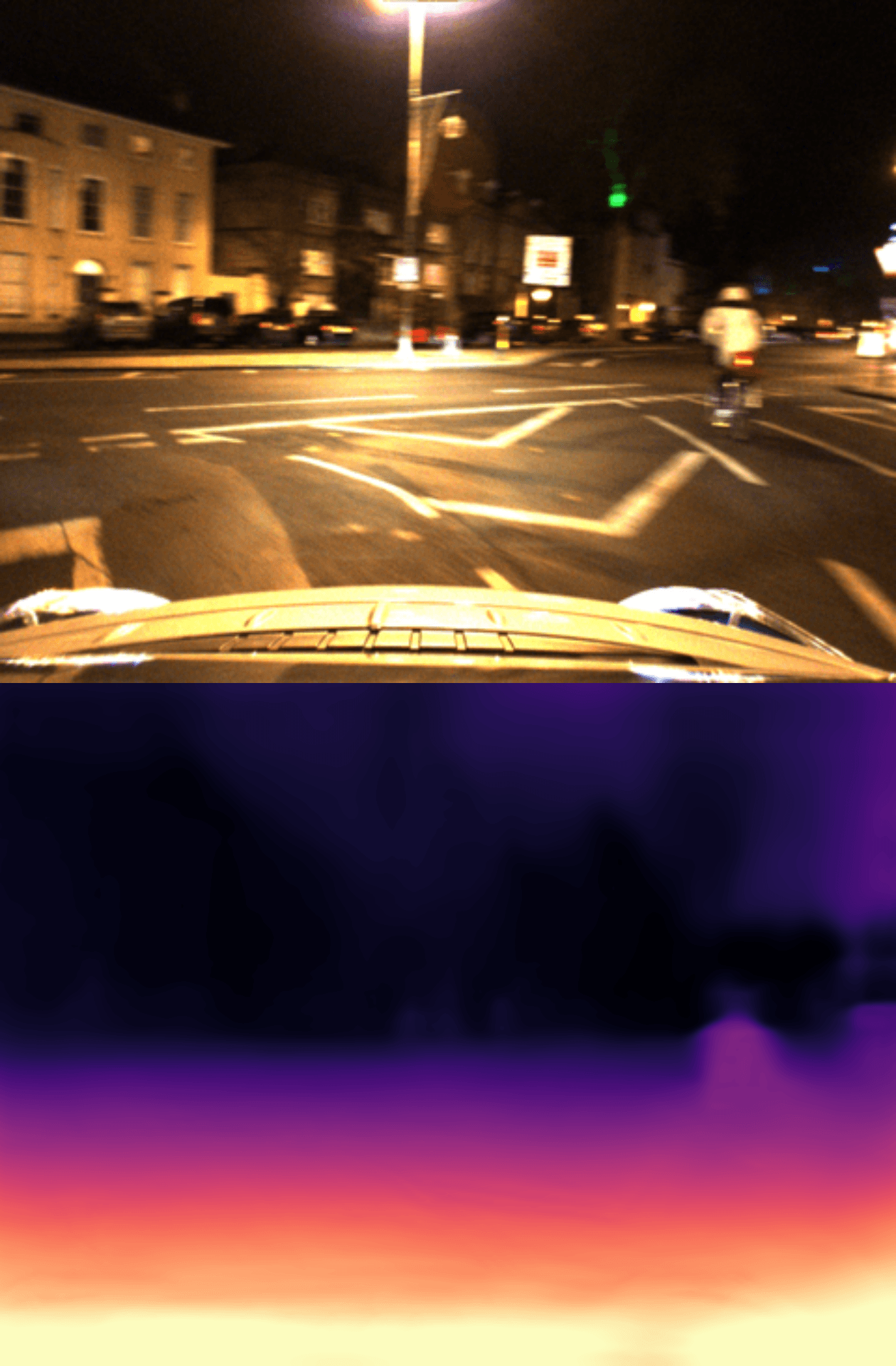}
    &\includegraphics[width=0.24\linewidth,trim={0 3cm 0 35cm},clip]{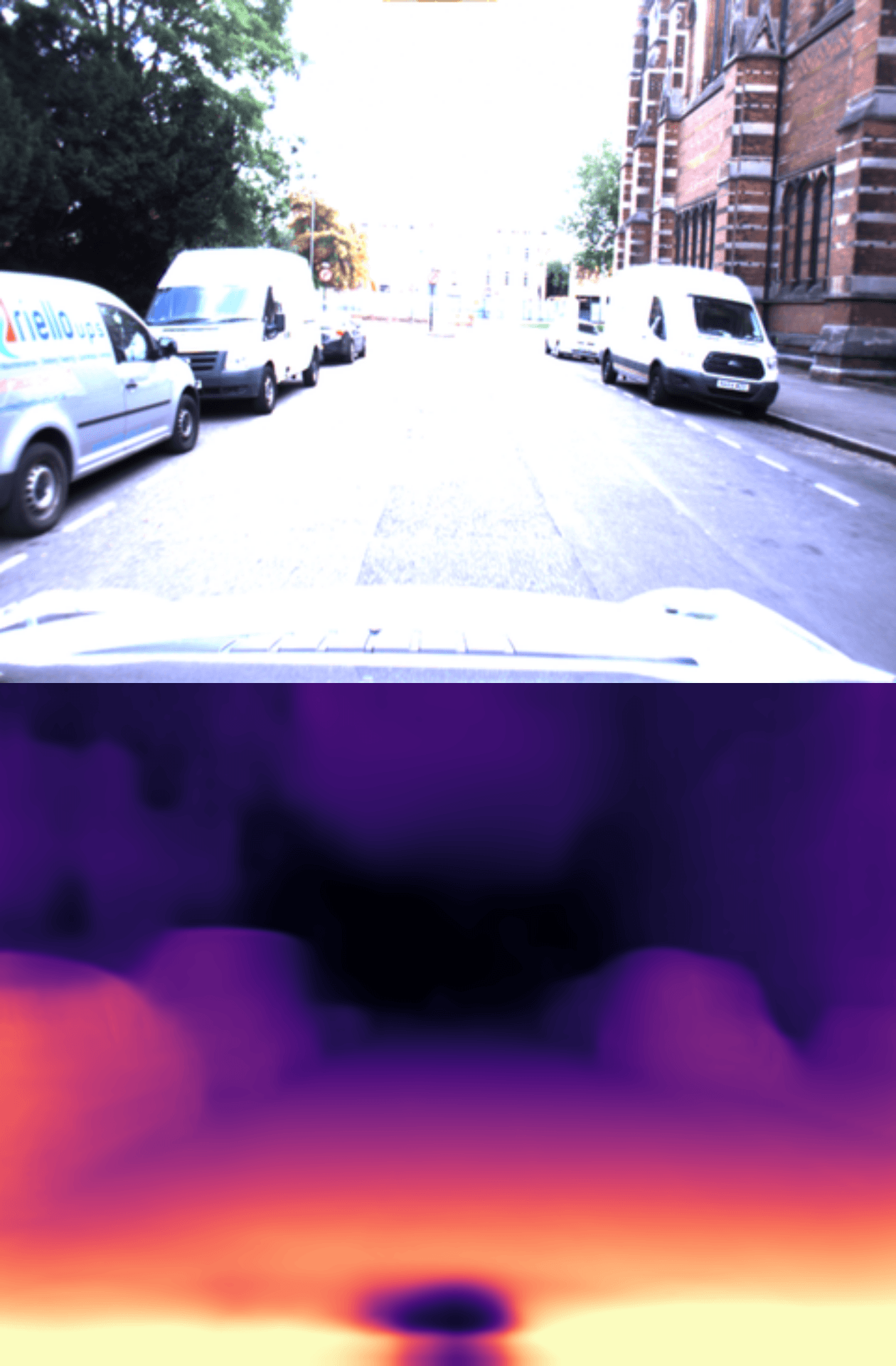}
    &\includegraphics[width=0.24\linewidth,trim={0 3cm 0 35cm},clip]{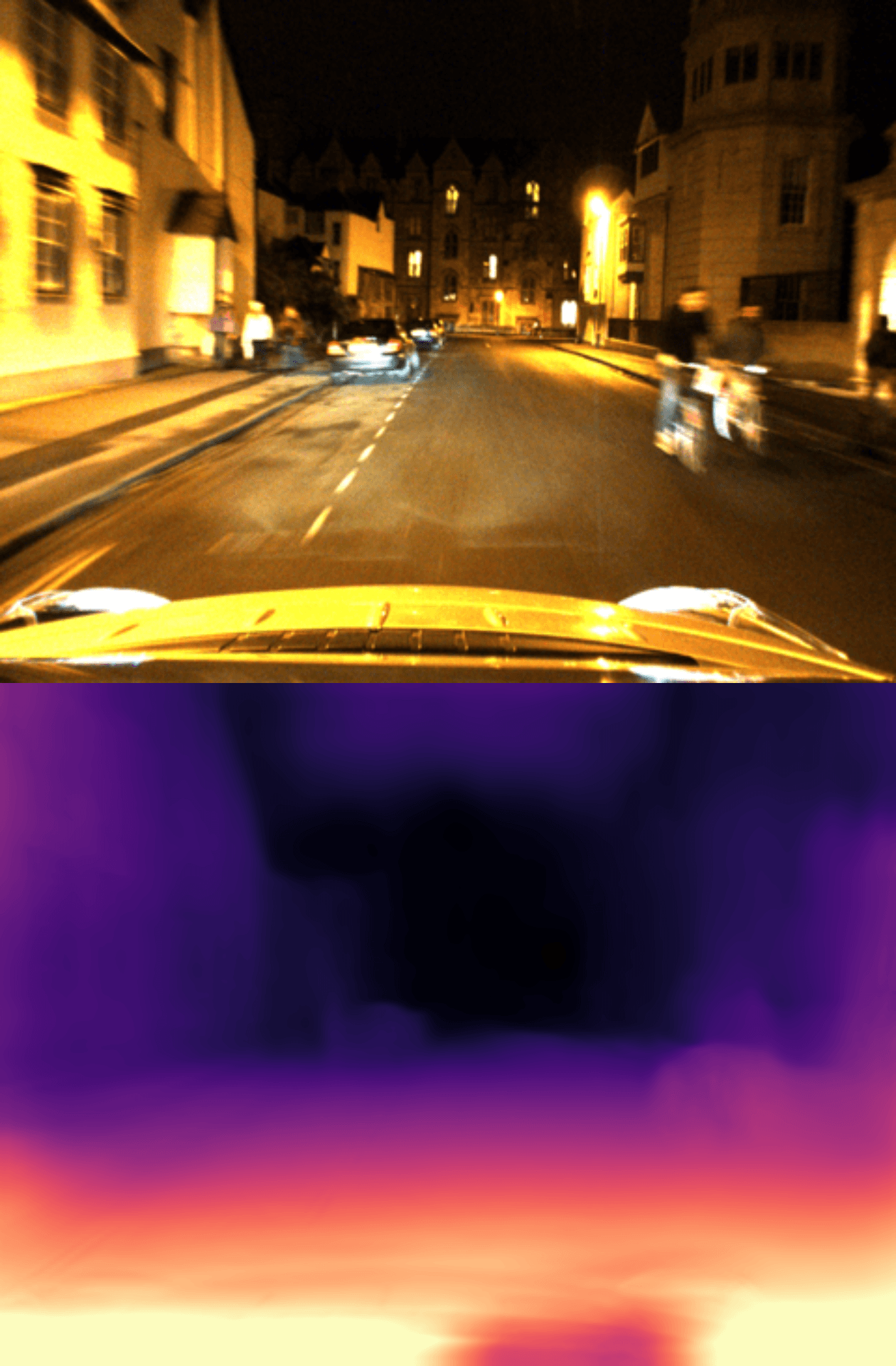}
    \\
    \end{tabular}
    \caption{Top: input images from the RobotCar dataset. Middle: estimated depth maps from Monodepth V2 \cite{Godard2019}. Bottom: estimated depth maps from \textit{DeFeat-Net}.}
    \vspace*{-0.3cm}
    \label{fig:robotcar_depth}
\end{figure*}

\begin{figure*}[!t]
	\vspace*{-1cm}
    \centering
    \includegraphics[width=0.33\linewidth]{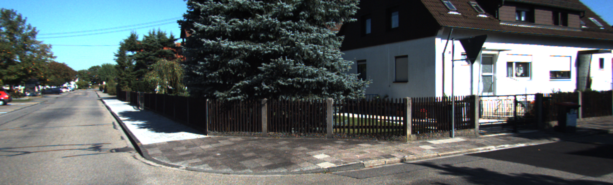}
    \includegraphics[width=0.33\linewidth]{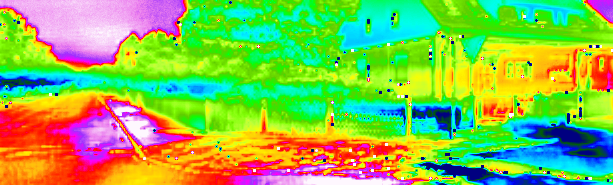}
    \includegraphics[width=0.33\linewidth]{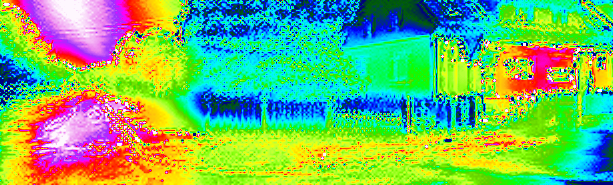}
    
    \includegraphics[width=0.33\linewidth]{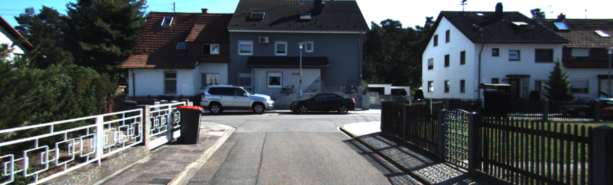}
    \includegraphics[width=0.33\linewidth]{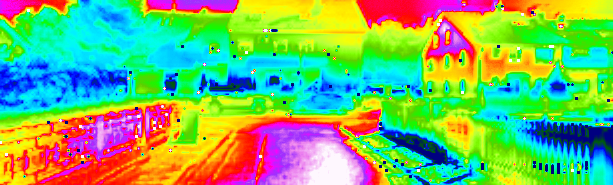}
    \includegraphics[width=0.33\linewidth]{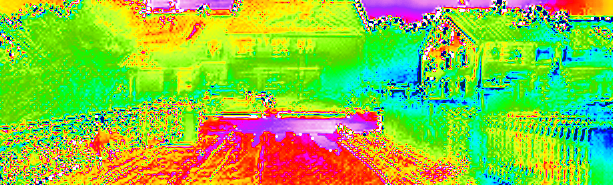}
    
    \includegraphics[width=0.33\linewidth,trim={0 50pt 0 0},clip]{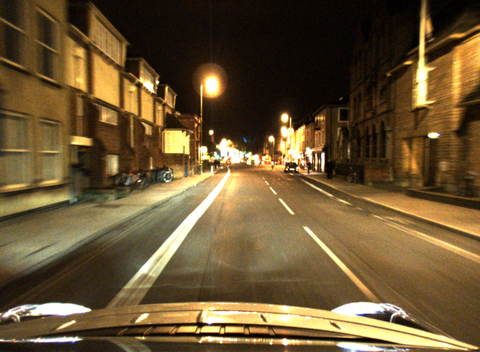}
    \includegraphics[width=0.33\linewidth,trim={0 35pt 0 0},clip]{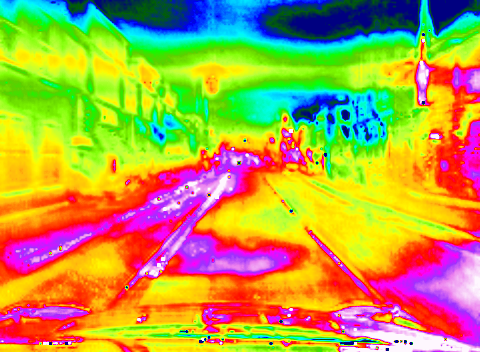}
    \includegraphics[width=0.33\linewidth,trim={0 35pt 0 0},clip]{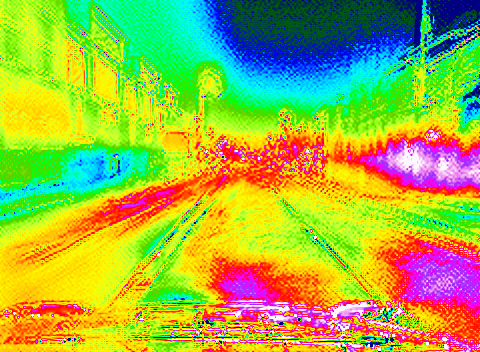}
    
    \includegraphics[width=0.33\linewidth,trim={0 50pt 0 0},clip]{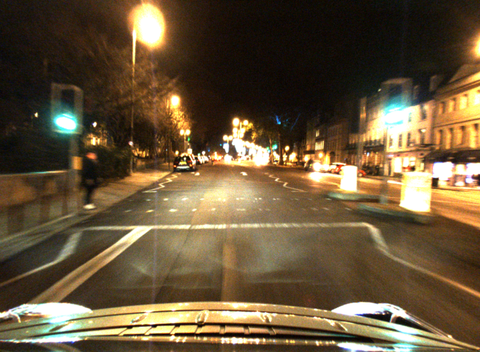}
    \includegraphics[width=0.33\linewidth,trim={0 35pt 0 0},clip]{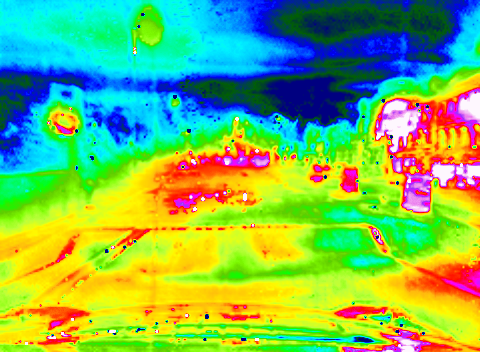}
    \includegraphics[width=0.33\linewidth,trim={0 35pt 0 0},clip]{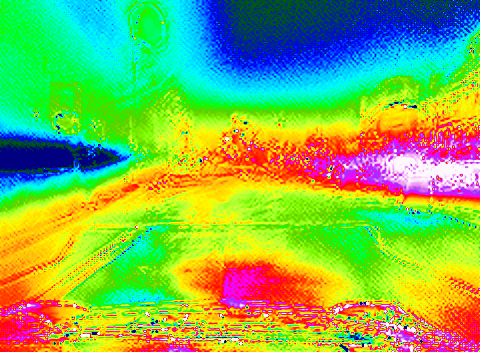}
    \caption{Feature space visualizations for \textit{DeFeat-Net} trained on the single-domain KITTI dataset (centre) and multi-domain RobotCar Seasons dataset (right).}
    \label{fig:robotcar_feat}
\end{figure*}

\subsection{Single Domain Evaluation}
We first evaluate our approach on the KITTI dataset, which covers only a single domain. 
For evaluation of depth accuracy, we use the standard KITTI evaluation metrics, namely the absolute relative depth error (ABS\_REL), the relative square error (SQ\_REL) and the root mean square error (RMSE). 
For these measures, a lower number is better. 
We also include the inlier ratio measures (A1, A2 and A3) of \cite{Godard2017} which measure the fraction of relative depth errors within 25\%, $25^2$\% and $25^3$\% of the ground truth. 
For these measures, a larger fraction is better.

To evaluate the quality of the learned feature representations, we follow the protocol of \cite{Spencer2019}. 
We compute the average distance in the feature space for the positive pairs from the ground-truth ($\mu_+$), and the negative pairs ($\mu_-$). 
Naturally a smaller distance between positive pairs, and a larger distance between negative pairs, is best. 
We also compute the Area Under the Curve (AUC) which can be interpreted as the probability that a randomly chosen negative sample will have a larger distance than the corresponding positive ground truth match. 
Therefore, higher numbers are better. 
Following \cite{Spencer2019} all three errors are split into both local (within 25 pixels) and global measurements.

The results of the depth evaluation are shown in Table~\ref{tab:kitti_depth} and the feature evaluation is shown in Table~\ref{tab:kitti_feat}. 
We can see that in this single-domain scenario, the performance of our technique is competitive with MonodepthV2 and clearly outperforms most other state-of-the-art techniques for monocular depth estimation. 
The results for \cite{Godard2019} were obtained by training a network using the code provided by the authors. 

Regarding the features, L2 denotes the L2-normalized versions, whereas G, L \& GL represent the different negative mining variants from \cite{Spencer2019}. 
We can also see that despite being unsupervised, our learned feature space is competitive with contemporary supervised feature learning techniques and greatly outperforms pretrained features when evaluating locally.
It is interesting, however, to note that the simple act of L2-normalizing can improve the global performance of the pretrained features.

Our feature space tends to perform better in the Global evaluation metrics than the local ones. 
This is unsurprising as the negative samples for the contrastive loss in (\ref{eq:contrastive}) are obtained globally across the entire image.

\subsection{Multi-Domain Evaluation}
However, performance in the more challenging RobotCar Seasons dataset demonstrates the real strength of jointly learning both depth and feature representations.
RobotCar Seasons covers multiple domains, where traditional photometric based monocular depth algorithms struggle and where a lack of cross-domain ground-truth has historically made feature learning a challenge. 
For this evaluation, we select the best competing approach from Table~\ref{tab:kitti_depth} (MonodepthV2) and retrain both it and DeFeat-Net on the RobotCar Seasons dataset. 
All techniques are trained from scratch.

The results are shown in Table~\ref{tab:robotcar_depth} and example depth map comparisons are shown in Figure~\ref{fig:robotcar_depth}. 
We can see that in this more challenging task, the proposed approach outperforms the previous state of the art technique across all error measures. 
While for the daytime scenario, the improvements are modest, on the nighttime data there is a significant improvement with around 10\% reduction in all error measures.

We believe that the main reason behind this difference is that in well-lit conditions, the photometric loss is already a good supervision signal.  
In this case, incorporating the feature learning adds to the complexity of the task.
However, nighttime scenarios make photometric matching less discriminative, leading to weaker supervision.
Feature learning provides the much needed invariance and robustness to the loss, leading to the significant increase in performance.

It is interesting to note that the proposed approach is especially robust with regards to the number of estimated outliers. 
The A1, A2 and A3 error measures are fairly consistent between the day and night scenarios for the proposed technique. 
This indicates that even in areas of uncertain depth (due to under-exposure and over-saturation), the proposed technique fails gracefully rather than producing catastrophically incorrect estimates.

Since previous state-of-the-art representations cannot be trained unsupervised, and RobotCar Seasons does not provide any ground-truth depth, it is not possible to repeat the feature comparison from Table~\ref{tab:kitti_feat} in the multi-domain scenario. 
Instead Figure~\ref{fig:robotcar_feat} compares qualitative examples of the learned feature spaces. 
For these visualizations, we find the linear projection that best shows the correlation between the feature map and the images and map it to the RGB color cube.
This dimensionality reduction removes a significant amount of discriminative power from the descriptors, but allows for some form of visualization.

In all cases, the feature descriptors can clearly distinguish scene structures such as the road. 
It is interesting to note that a significant degree of context has been encoded in the features, and they are capable of easily distinguishing a patch in the middle of the road, from one on the left or right, and from a patch of similarly colored pavement. 
The feature maps trained on the single domain KITTI dataset can sometimes display more contrast than those trained on RobotCar Seasons. 
Although this implies a greater degree of discrimination between different image regions, this is likely because the latter representation can cover a much broader range of appearances from other domains. 
Regarding, the nighttime features, it is interesting that those trained on a single domain seem to exhibit strange behaviour around external light sources such as the lampposts, traffic lights and headlights. 
This is likely due to the bias in the training data, with overall brighter image content.

\begin{table*}[!t]
\centering
\begin{tabular}{c|c|ccccccc}
Dataset & Method & Abs-{Rel} & Sq-Rel & RMSE & RMSE-log & A1 & A2 & A3 \\ \hline
KITTI & DeFeat (no feat) & \textbf{0.123} & 0.948 & 5.130 & \textbf{0.197} & \textbf{0.863} & \textbf{0.956} & \textbf{0.980} \\ 
KITTI & \textbf{DeFeat} & 0.126 & \textbf{0.925} & \textbf{5.035} & 0.200 & 0.862 & 0.954 & \textbf{0.980} \\ \hline
RobotCar Day & DeFeat (no feat) & 0.274 & 3.885 & \textbf{8.953} & 0.335 & \textbf{0.640} & \textbf{0.853} & 0.934 \\
RobotCar Day & \textbf{DeFeat} & \textbf{0.265} & \textbf{3.129} & 8.954 & \textbf{0.323} & 0.597 & 0.843 & \textbf{0.935} \\\hline
RobotCar Night & DeFeat (no feat) & 0.748 & 13.502 & \textbf{8.956} & 0.657 & 0.393 & 0.624 & 0.759 \\
RobotCar Night & \textbf{DeFeat}  & \textbf{0.335} & \textbf{4.339} & 9.111 & \textbf{0.389} & \textbf{0.603} & \textbf{0.828} & \textbf{0.914} \\
\end{tabular}
\caption{Performance with and without concurrent feature learning, on each dataset}
\vspace*{-0.4cm}
\label{tab:ablation}
\end{table*}

\subsection{Ablation}
Finally, for each dataset we explore the benefits of concurrent feature learning, by re-training with the FeatNet subsystem disabled. 
As shown in Table~\ref{tab:ablation}, the removal of the concurrent feature learning from our technique causes a small and inconsistent change on the KITTI and RobotCar Day data. 
However, on the RobotCar Night data, our full approach drastically outperforms the version which does not learn a specialist matching representation. 
For many error measures, the performance doubles in these challenging scenarios, and the reduction in outliers causes a three-fold reduction in the Sq-Rel error.

These findings reinforce the observation that the frequently used photometric warping loss is insufficient for estimating depth in challenging real-world domains.

\section{Conclusions \& Future Work}
This paper proposed DeFeat-Net, a unified framework for learning robust monocular depth estimation and dense feature representations. 
Unlike previous techniques, the system is able to function over a wide range of appearance domains, and can perform feature representation learning with no explicit ground truth. 
This idea of co-training an unsupervised feature representations has potential applications in many areas of computer vision beyond monocular depth estimation.

The main limitation of the current approach is that there is no way to enforce feature consistency across seasons. 
Although depth estimation and feature matching work robustly within any given season, it is currently unclear weather feature matching between different seasons is possible. 
It would be interesting in the future to explore cross-domain consistency as an additional training constraint. 
However, this will necessitate the collection of new datasets with cross seasonal alignments.

\subsection*{Acknowledgements}
\vspace{-0.3cm}
This work was partially funded by the EPSRC under grant agreements (EP/R512217/1, EP/S016317/1 and EP/S035761/1). 
We would also like to thank NVIDIA Corporation for their Titan Xp GPU grant.
\vspace{-0.4cm}

{
\small
\bibliographystyle{ieee_fullname}
\bibliography{CVPR2020_2}
}
\end{document}